# Sixteen models, fewer than two voices: measuring ensemble dispersion where no answer is uniquely correct

Mario Vega-Barbas, PhD[1]; Lidia Mora-Valenciano[2]; Iván Pau, PhD[1]; Fernando Seoane, PhD[3,4,5,6]; Farhad Abtahi, PhD[3,4,7]

[1] ETSIS de Telecomunicación, Universidad Politécnica de Madrid, Madrid, Madrid, Spain
[2] Independent Researcher, Guadalajara, Spain
[3] Department of Clinical Science, Intervention and Technology, Karolinska Institutet, Huddinge, Stockholm, Sweden
[4] Department of Clinical Physiology, Karolinska University Hospital, Huddinge, Stockholm, Sweden
[5] Department of Medical Technology, Karolinska University Hospital, Stockholm, Sweden
[6] Department of Textile Technology, University of Borås, Borås, Västra Götaland, Sweden
[7] Department of Biomedical Engineering and Health Systems, KTH Royal Institute of Technology, Huddinge, Stockholm, Sweden

**Corresponding Author:**
Mario Vega-Barbas, PhD
ETSIS de Telecomunicación
Universidad Politécnica de Madrid
Nikola Tesla, s/n (Campus Sur)
Madrid, 28031, Spain
Phone: +34 669 971 639
Email: mario.vega@upm.es

*Body word count: 11622 (excluding abstract, headings, figure captions and references). Figures: 3. Tables: 0 (seven tables are in the Supplementary Material, sections S1 to S11).*

## Abstract

Sixteen language models drawn from ten families produced, on average, the semantic diversity of 1.69 distinct formulations of a psychotherapeutic case, against a single-model baseline of 1.43 from resampling one model's own runs. Ensembles place more than one reading before a decision-maker on the premise that several models supply several perspectives. Dispersion over their outputs is measured both as diversity and as uncertainty, and both traditions validate it against a correctness criterion that this task does not admit. Measuring diversity is a solved problem: the Vendi Score, the exponential of the von Neumann entropy of a similarity matrix, is an effective number of distinct elements, intensive and robust to redundancy. What a single aggregate does not say is where the diversity comes from. We define a per-model dissent contribution, the complement of a model's mean similarity to the other members of its ensemble: a magnitude from the same matrix, not a decomposition of the spectral index, whose maximum identifies the most divergent voice. Crossing model and case, we test as a preregistered hypothesis whether model identity accounts for a non-zero share of the variance in dissent, and characterise the structure that test detects. The panel formulated fifteen stratified psychotherapeutic vignettes, yielding 7,082 formulations for analysis. Model identity was a detectable structuring factor of the dissent that remained, but the usual categories recovered it only partly: scale differences pointed in opposite directions across pairs, family grouped models on only five two-member lines, and the most divergent voice changed with panel composition, so that the surfaced outlier describes the ensemble rather than the model. Dissent did not track the interpretive openness for which the case bank was stratified; it was organised by clinical content instead, leaving the dispersion an ensemble produces a property to measure rather than assume.

## Introduction

Ensembles of language models are increasingly used to place more than one reading of a problem in front of a decision-maker, in arrangements ranging from independent panels to interacting multi-agent systems. The premise of that arrangement is that several models supply several perspectives, and it is a premise that is rarely checked. What is missing is a measurement of how much genuinely distinct perspective an ensemble produces on a given case, as opposed to how many models it contains, and of what structures that quantity: whether it is a property of the models assembled, of the case they are given, or of the composition of the panel itself.

Several things turn on the answer. The design of an ensemble does: if the number of models is a poor proxy for the number of perspectives, then adding members is not in itself a way of widening the space of readings offered, and the effective count is the quantity a designer needs. Meaningful human oversight of high-risk systems is another. Regulation requires it (European Parliament and Council of the European Union, 2024) without specifying when it should be exercised; how to exercise it so that it is meaningful is an active area of review (Lazaros et al., 2026); and placing a clinician in the loop does not by itself make the oversight meaningful (Toro-Tobon et al., 2026). Whether a system has converged on a single reading is information a supervisor might want, but establishing that it is the right criterion for allocating expert attention is a question about people and their decisions, and it is not the question answered here. What can be settled first is what a measure of dispersion over an ensemble's output registers, and what it does not.

The distinction matters because the two quantities come apart. Language models trained on overlapping corpora and aligned towards overlapping objectives produce texts that differ in wording while converging in substance. This homogeneity has been documented directly across models (Wenger and Kenett, 2025), traced to the composition of post-training data rather than to inference-time settings (Karouzos et al., 2026), and observed in multi-agent arrangements, where added agents yield diminishing marginal diversity and interaction amplifies shared priors rather than introducing variety (Chen et al., 2026). The convergence is not confined to wording: across large panels of models, the errors of more accurate systems are more correlated with one another than those of weaker ones, and this holds across architectures and providers (Kim et al., 2025; Goel et al., 2025), so heterogeneity of provenance is a weaker guarantee of independence than it appears. Counting models is therefore a poor proxy for counting perspectives, and any argument that rests on the number of members assumes what needs to be measured.

Quantifying the latter is a solved problem in its general form. The Vendi Score (Friedman and Dieng, 2022) defines the diversity of a collection as the exponential of the von Neumann entropy of a normalised similarity matrix over its elements, yielding an effective number of distinct items that is bounded between one and the size of the collection. The construction has two properties that this purpose requires and that simpler alternatives lack. It is intensive, so that a value carries the same meaning whatever the size of the panel and a threshold set for one configuration remains interpretable in another. And it is robust to redundancy, so that two near-identical members are counted as approximately one rather than mistaken for consensus or, in determinant-based alternatives, for numerical pathology. The measure has been applied to the outputs of language models in several settings, including educational item generation, scientific ideation, and the study of diversity collapse in multi-agent systems.

The same functional has been reached independently from a second direction. Work on uncertainty quantification in language models measures the semantic dispersion of the texts a model produces for a single input, first by clustering generations into classes of equivalent meaning and taking the entropy of the cluster distribution (Kuhn et al., 2023; Farquhar et al., 2024), and subsequently by replacing that hard clustering with a positive semidefinite unit trace kernel over the generations and taking its von Neumann entropy (Nikitin et al., 2024), a construction shown to generalise the clustering version and coinciding with the index used here up to the choice of kernel. That two literatures converge on one quantity, one asking how diverse a collection is and the other how uncertain a model is, is a reason to take the measure seriously rather than a reason to prefer either name for it.

What those literatures share, and what separates them from the present study, is the criterion against which dispersion is validated. In the ensemble tradition it is assessed by whether it predicts the accuracy a vote gains over its best member, which requires labels (Kuncheva and Whitaker, 2003; Kim, 2026). In the uncertainty tradition it is assessed by whether it separates correct generations from incorrect ones, which requires a fact of the matter (Farquhar et al., 2024). This holds when the comparison is made across models rather than within one: the recent extension of that tradition to heterogeneous panels reports its estimator to be most informative on tasks with a unique correct answer, and identifies tasks admitting several distinct correct responses as the case in which disagreement no longer reflects uncertainty (Hamidieh et al., 2026). Case formulation is such a task. The same presentation admits several defensible readings (Hider and Sim, 2024), no correctness criterion is available, and none is used here. What a measure of dispersion registers under that condition has not been characterised, and characterising it is the object of this study.

What has not been established is whether such a measure supports the inferences that reasoning about an ensemble requires. Existing applications report a single aggregate figure per collection, which answers how much diversity there was but not where it came from. For ensemble design, and for anyone reading what an ensemble produces, the more useful questions are about structure: whether the identity of a model systematically determines how far its contribution falls from the consensus, whether that structure can be predicted from the properties by which models are ordinarily grouped, and whether the member that diverges most is a stable property of that model or an artefact of the panel it happens to sit in. Answering these requires decomposing the aggregate rather than reporting it alone.

We therefore define a per-model dissent contribution: for each member, the complement of its mean similarity to the remaining members. Its maximum identifies the most divergent voice, and its average across members is the panel's mean pairwise dissent, a standard summary of internal diversity that the Vendi Score's reference implementation also computes. Attributing to each element of a collection a share of the collection's diversity is likewise not new: derivative work on the Vendi Score assigns every element a weight for how much it contributes to the whole, and uses those weights to reweight or select elements (Pasarkar and Dieng, 2025). Our use of the per-element quantity is different. We treat the per-model contribution as a response variable over which model identity and case identity are crossed as factors, and estimate the share of its variance attributable to each, so that model-level structure can be tested rather than elements ranked. We use it to test, as a preregistered hypothesis, whether model identity accounts for a non-zero share of the variance in dissent, and to characterise the structure that test detects.

The characterisation is performed on a panel of sixteen models spanning ten families and, within four of them, two scales, evaluated on a fixed bank of stratified psychotherapeutic vignettes. Case formulation in psychotherapy was chosen as the domain because its structural interpretive ambiguity elicits genuine disagreement: the same presentation admits several defensible readings (Hider and Sim, 2024), and a clinician choosing among them exercises judgement rather than retrieving a fact. That property makes it a demanding setting for a dissent measure: if an ensemble is going to diverge anywhere, it is on material of this kind. The bank is stratified on two crossed dimensions, the clinical picture and the interpretive openness of the case, the second of which affords a check on what the measure registers. The bank was held fixed while the panel varied, so that variation in measured dissent is attributable to the models. The hypothesis, the definition of the measure, the composition of the panel and the analysis plan were registered publicly before any data were generated.

Four findings follow. The panel of sixteen behaved, on average, as the equivalent of fewer than two distinct formulations, against a single-model baseline of 1.43 obtained from one model's own stochastic variation, extending to a heterogeneous multi-model panel in an interpretive clinical domain the convergence reported elsewhere for creative and open-ended tasks. Model identity is nonetheless a detectable structuring factor of the dissent that remains, which was the registered prediction and is supported. That structure is only partly recovered from the categories ordinarily used to reason about models: within-family scale differences are individually clear but point in opposite directions across pairs, family membership groups models but only on the thin evidence of five two-member lines, and the identity of the most divergent voice changes when the composition of the panel changes, so that an outlier surfaced to a reader is a statement about the ensemble rather than about the model surfaced. And the dissent the panel produces is organised by the clinical content of the case rather than by the interpretive openness for which the material was stratified, which is the first characterisation of what the quantity registers where correctness does not apply.

## Materials and Methods

### Preregistration

The study was preregistered before any data were generated. The registration fixes the hypothesis, the definition of the index, the composition of the panel, the design, and the analysis plan, and is available at osf.io/c5qk7, sealed on 21 July 2026. Three departures from the registered protocol were adopted: two before any data were generated, both arising from provider-level constraints identified in a verification carried out in advance of execution, and a third after execution, a mechanical validity filter applied to the formulations obtained. They are described in the relevant subsections below and recorded in full in the accompanying deviations document. No analysis was performed until data collection was complete, and the confirmatory analysis was executed before any exploratory analysis.

### Case material

The case material is a bank of fifteen synthetic clinical vignettes, written by the study's clinical co-author, an experienced psychotherapist, with methodological coordination by the lead author, and constructed so that the person who wrote the vignettes took no part in their validation. The vignettes are stratified along two crossed clinical dimensions. The first is the clinical picture, in four categories, depression, trauma, substance abuse, and relational conflict, distributed as four, four, three, and four vignettes. The second is the interpretive openness of the case, in three categories, a shared framework that reads the case ineffectively, a shared framework that reads it effectively, and a case that genuinely admits distinct framings, distributed as six, two, and seven vignettes. The two dimensions cross into twelve possible cells, of which ten are populated and two, documented as such, are empty. Each vignette runs to roughly one hundred and fifty to two hundred words of continuous prose in Spanish and carries no diagnosis, aetiology, or treatment plan, leaving the framing to the models.

The ten therapeutic frameworks offered to the models are the cognitive-behavioural, the humanistic-existential, the psychodynamic-psychoanalytic, the integrative, the mindfulness-based or third-generation, the attachment, the somatic or sensorimotor trauma, the structural-strategic systemic, the transgenerational Bowenian systemic, and the narrative frameworks.

The bank was validated by a panel of four practising therapists, blind and independent of one another. For each vignette the therapists rated two independent dimensions on five-point Likert scales, the clinical plausibility of the vignette and its clinical sufficiency, and rated the representativeness of the bank as a whole in a closing block. On the descriptive metrics that served as the primary reading the bank scored as acceptable, with a median plausibility rating of 4.5 and a median sufficiency rating of 3.5, every vignette meeting the minimum acceptability threshold on plausibility and all but two on sufficiency. Inter-rater agreement measured by the intraclass correlation coefficient, in its two-way random, absolute-agreement, average-measures form (Shrout and Fleiss, 1979; McGraw and Wong, 1996), was low, at 0.033 for plausibility and −0.127 for sufficiency, both in the poor range (Koo and Li, 2016) and with wide confidence intervals. The four raters were drawn from a single clinical practice of six therapists, none of them a member of the research team, and were selected from that pool so as to differ from one another in clinical orientation rather than to be alike; the heterogeneity of the panel was therefore intended, within the limits of a convenience pool. Raters differing in theoretical background and in tolerance for interpretive ambiguity produce conservative agreement coefficients, and with fifteen vignettes and four raters the confidence intervals on the coefficient are necessarily wide. The exercise is accordingly read as establishing the plausibility and representativeness of the bank as characterisation material, on the descriptive acceptability metrics reported above, rather than as a psychometric validation of it. This limitation is returned to in the discussion.

The bank was held fixed throughout. Because the stimuli were constant while the models varied, variation in measured dissent is attributable to the models rather than to the material, which is the condition a characterisation of model-level structure requires. The bank was deliberately not expanded to a larger random set: a stratified and clinically validated bank carries more design information than a larger unstructured sample, and the statistical power of the design derives from the number of runs and the crossed structure rather than from the number of vignettes.

**Panel of models**

Sixteen large language models were queried through a single provider gateway, spanning ten model families and including four within-family scale pairs: Claude Haiku 4.5 and Claude Sonnet 4.5, Ministral 14B and Mistral Large, Gemma 4 26B and Gemma 4 31B, and DeepSeek V4 Flash and DeepSeek V4 Pro. The panel was composed to permit both family-level and within-family scale comparison, following the multi-scale panel construction used in MEDLEY-BENCH (Abtahi et al., 2026). Model identifiers were pinned to dated versions resolved against the provider catalogue, so that the executed panel refers to specific model snapshots rather than to floating aliases; the complete pinned list is given in the supplementary material. Free-tier variants were excluded, since their throughput limits and fluctuating endpoints are incompatible with a reproducible run of this size. Two levels of grouping are distinguished throughout. The family follows the provider taxonomy, under which the two Gemini generations and the two Gemma models all belong to Google. The line is the finer division within it that separates Gemini from Gemma. For every model not from Google the two coincide, so the sixteen models fall into ten families and eleven lines.

The registration specified a panel of seventeen models. One of them was excluded before execution because it emits reasoning tokens by design and does not accept their deactivation through any available provider, which would have placed that model under conditions different from the other sixteen. Since the study characterises structure attributable to model identity, an asymmetry of treatment within the panel would have confounded the quantity of interest, and the model was removed. It belonged to no scale pair, so its exclusion left the four within-family comparisons intact.

**Generation procedure**

Each model received the same instruction template and produced an independent formulation of each case, in an isolated context with no access to the outputs of the other models. Generation used a temperature of 0.4 and an output limit of 4096 tokens. Reasoning was disabled where the provider permits it, so that the output limit would act as an effective bound on the length of the formulation, as it did in the earlier three-model exercise whose parameters the registration undertook to hold constant. One model attaches a reasoning field to its response regardless of the setting; that field carries no billed reasoning tokens and is not processed by the pipeline, which embeds the response content only, so the model was used as delivered.

The registration specified an output limit of 2048 tokens. Verification before execution established that this value truncated the formulations of two models, which terminated on length rather than on completion, and that truncated formulations would have contaminated the measurement of semantic dissent with artefacts of the cut. The limit was raised to 4096, a value verified not to alter the behaviour of the models that already completed within the original bound. Forty-seven of the seven thousand one hundred and sixty formulations obtained (0.66 per cent) still reached the raised limit.

No fusion step was applied. Because the object of measurement is the dispersion among the formulations rather than an integrated consensus response, no fused output was produced, which removes a source of variance extraneous to the dissent signal.

The instruction template casts the model as an experienced clinical supervisor writing to a supervisee and asks it for a deliberative reading of the presented case rather than for operational instructions, requiring a committed and recognisable clinical voice rather than a neutral consensus answer. The template is written in Spanish, the language of the bank, and the formulations were produced in Spanish accordingly. The response follows a fixed structure that opens with a single line declaring the therapeutic framework from which the reading is articulated, continues with the clinical formulation as the bulk of the text, then a section on what the supervisee should explore in subsequent sessions, and closes with a section on the zones of uncertainty, within an overall length of six hundred to eight hundred words. The ten frameworks are presented as a list of options from which the model selects the one it finds most informative for the case at hand, declaring a combination where it reads through more than one. The template is held byte-identical across all runs except for the ordering of that list of ten frameworks, the only element rewritten from run to run; rewriting the list in its canonical order reproduces the template exactly, a property verified before execution. The template was held identical to the one used in the earlier three-model exercise, and in consequence its text still refers to a system comparing the readings of several supervisors, a description that predates the enlargement of the panel and has no bearing on how each formulation is generated in isolation.

## Design

Presentation order of the ten therapeutic frameworks was counterbalanced with a Williams Latin square (Williams, 1949) of order ten, which yields ten unique presentation sequences; each sequence was replicated three times, giving thirty runs per vignette and, over fifteen vignettes, four hundred and fifty ensemble runs per model. Across the panel this yields seven thousand two hundred model-level formulations. The counterbalancing controls for the effect of the order in which frameworks are offered, an ordering bias documented for options presented to language models (Lin et al., 2024); the same set of orders was applied to every vignette and every model, so that no model encountered a systematically different sequence. The design is fully crossed and within-panel: every model was observed on every case, with both factors crossed rather than assigned between subjects.

The order plan was generated under a fixed seed of 42 and records the identifying hashes of the two frozen inputs against which it was built: the SHA-256 hash of the prompt template, 596518f4018ea981669f756b9002245f709898adab0ea49c37afc30a1270b396, and the SHA-256 hash of the vignette bank, ffd5d46b1c0f3a7efc105ab58ce22cc0c2c38c20d5eae03206270b2e0b52c1c2. Any run of the study can therefore be traced to the exact prompt and bank from which it was produced.

## The dissent index

For each run, the formulations produced by the panel were embedded with a single embedding model at 3072 dimensions and normalised to unit length, and their Gram matrix M was formed, which is the matrix of cosine similarities among the formulations. Each formulation was truncated to eight thousand characters before embedding to remain within the model's input limit; in a single run one response still exceeded the limit and that run's responses were truncated to six thousand characters, with all voices retained. M is symmetric and positive semidefinite with unit diagonal, so its trace equals the number of voices.

The dissent of the ensemble on a case is quantified by the Vendi Score (Friedman and Dieng, 2022), which defines the diversity of a collection as the exponential of the von Neumann entropy of a normalised similarity matrix over its elements. The eigenvalues of M, normalised by the trace, define a probability distribution, equivalent to the eigenvalues of the density matrix obtained by dividing M by its trace. The entropy of that distribution is S equal to minus the sum of p_i times the natural log of p_i, with the convention that zero times the logarithm of zero is zero, and its range is zero to log n. We report both the entropy normalised by log n, which we denote S_norm and which is bounded in the unit interval, and its exponential, which is the Vendi Score itself and which we refer to as the effective number of voices. The unnormalised entropy in nats is reported in the supplementary material.

Two properties of this construction make it suitable for the present purpose, and rule out simpler alternatives. It is intensive: normalisation anchors the maximum independently of panel size, so that a value carries the same meaning whatever the number of models and a threshold set for one configuration remains interpretable in another. And it is robust to redundancy: two near-identical voices share an eigenvalue and contribute a term that tends smoothly to zero, whereas a geometric mean of eigenvalues is dragged by small eigenvalues and confounds redundancy with consensus. Both properties are necessary for a heterogeneous panel that includes models of the same family at different scales. The entropy is computed over the eigenvalues of M and not over its entries: measuring dissent as the Shannon entropy of the raw off-diagonal similarities degenerates, because normalising high and mutually close similarities saturates the entropy near its maximum irrespective of the case, whereas the eigenvalues retain a discriminating structure.

The effective number of voices expresses how many genuinely independent contributions the panel is equivalent to producing on a given case once redundancy between models is discounted, and is bounded between one and the number of voices. It and the normalised index are two scales of the same quantity; the normalised index is used for the statistical analysis and the effective number of voices for interpretation. Both are reported as continuous quantities, without discretisation into bands: deriving bands from the same material subsequently analysed would be circular, and no bands were registered.

The per-model dissent contribution is defined, for each voice, as one minus its mean similarity to the remaining voices of the same run. It is the additive per-voice decomposition of the ensemble's mean pairwise dissent, not of the spectral index, since the mean of the contributions across voices equals the mean off-diagonal dissent, and its maximum identifies the voice separating most from the consensus. This mean off-diagonal dissent is proportional, by a factor of $n/(n-1)$, to the internal diversity that the Vendi Score's reference implementation computes, so the quantity is not new to this work. It is the response variable of the confirmatory analysis.

Eigenvalues were obtained by symmetric eigendecomposition. Negative eigenvalues arising from rounding were set to zero rather than taken in absolute value, since a rounding-induced negative eigenvalue represents a direction of no variance; eigenvalues below a relative tolerance of $10^{-12}$ times the largest were excluded from the entropy sum; normalisation was by the trace. The implementation was verified against the limiting cases before execution, reproducing an index of zero and one effective voice for identical formulations and an index of one and n effective voices for mutually orthogonal ones.

**Handling of incomplete runs**

Where a model returned no usable formulation for a given run, the call was retried up to four times. Where a voice remained absent after retries, the similarity matrix and the indices were computed from the available voices, with the reduced voice count and the identity of the absent model recorded for that run. This rule was fixed in the design documents before any data were generated and applied as specified; it is not part of the sealed registration, and the deviations record states so. Forty of the four hundred and fifty runs were computed from fifteen voices rather than sixteen, the absent voice belonging to one model in thirty-eight of them and to another in two; the absences fell in a contiguous window of the run sequence and are consistent with a transient provider instability rather than with any property of the cases. The absences are tabulated by model and vignette in the supplementary material. No post-hoc recovery of absent voices was attempted, since re-running only the runs known to have failed would apply an unregistered procedure selectively to the failures and would generate data in a different temporal window from the rest of the dataset.

**Validity of formulations**

Inspection of the distribution of per-model dissent contributions after execution revealed a small number of extreme values, far above the bulk of the distribution. Review of the formulations behind them showed that they were degenerate outputs rather than divergent readings: fragments left by a provider error, streams of a single repeated character, looping n-grams, or text in the wrong language. The validity check implemented during generation verified only that a response was non-empty, so these outputs passed it and were embedded and analysed as if usable, without triggering the retry rule that a failed formulation was meant to invoke.

To handle them uniformly, and after the fact, eight mechanical criteria were frozen before they were applied. Each is a content-level test with an exact threshold, and none uses the per-model dissent, the spectral index, the effective number of voices, or the identity of the model, so the filter is blind both to the magnitude of the dissent a formulation carries and to which model produced it. The criteria and their thresholds are recorded in the accompanying validity criteria document. One of the eight, a check for the framework declaration, was withdrawn on application because it flagged legitimate formulations that declared their framework in a variant form; the remaining seven were applied uniformly to the seven thousand one hundred and sixty formulations obtained. Seventy-eight failed, and each is treated as an absent voice under the same missing-voice rule that governs the provider absences, a rule adopted before any data were generated and documented in the deviations record. The failures are broken down by criterion and by model in the supplementary material.

Treating the degenerate formulations as absent removes artefactual variance rather than adding structural signal. Being nearly unrelated to the other formulations of their runs, the degenerate outputs carried extreme dissent contributions that inflated the within-run variance of the response. Their removal lowers the absolute model and case variance components, from $2.12 \times 10^{-4}$ to $1.56 \times 10^{-4}$ for the model factor and from $7.66 \times 10^{-5}$ to $6.58 \times 10^{-5}$ for the case factor, while the residual variance collapses from $1.44 \times 10^{-3}$ to $1.09 \times 10^{-4}$. The share of variance the structural factors explain rises accordingly, but this reflects the elimination of artefactual within-run noise, not an increase in the model effect, whose absolute magnitude is smaller after the filter than before it.

**Statistical analysis**

The confirmatory hypothesis was that model identity accounts for a share of the variance in the per-model dissent contribution greater than zero. A cross-classified mixed-effects model was fitted to the per-model contributions with model identity and case identity as crossed random factors and no fixed-effect predictors beyond the intercept, and the variance components attributable to each factor and to residual variation were estimated. The hypothesis was evaluated on the model-factor component against a Monte Carlo permutation reference distribution constructed by permuting model labels within cases, using 99,999 resamples, which places the floor of the p-value at $10^{-5}$, under a seed fixed in the analysis code before execution. The test is one-sided by construction, since a variance component is non-negative and the prediction is directional. A two-way fixed-effects decomposition of the same response was reported as a registered sensitivity analysis. Effect sizes for the fixed-effects analyses are reported as classical eta squared, the sum of squares of the effect over the total sum of squares of the response, and not as partial eta squared, whose denominator is the effect together with its own error (Levine and Hullett, 2002; Lakens, 2013). The distinction is consequential wherever a covariate is entered: with type-II sums of squares the terms of the table no longer add to the total once they are collinear, so a ratio taken against the table rather than against the total would change with the model being fitted and would not be comparable across fits. As a registered internal coherence check, the three reference cases of the earlier three-model pilot were recomputed under the spectral formulation and the ordering they received under the earlier determinant-based formulation was confirmed to be preserved; the values are given in S11 of the supplementary material. No correction for multiple comparisons applies, since there is a single confirmatory hypothesis.

Four analyses were registered in advance as exploratory: the partition of variance between the model and case factors, the effect of scale within families, the identity and stability of the most divergent voice at panel scale, and the grouping of models by family. None carries a registered directional prediction. In particular, no direction was registered for the partition of variance, because preliminary estimates from the earlier three-model exercise pointed towards the case factor accounting for more of the magnitude variance and registering a direction unsupported by the available evidence would have been inappropriate; and none was registered for the scale effect, because the nearest published evidence reports scale effects on multi-model behaviour to be dissociated and non-monotonic (Abtahi et al., 2026). The exploratory analyses are reported descriptively and were executed after the confirmatory analysis, in an order that places first the analysis most likely to complicate the confirmatory finding.

Uncertainty in the exploratory quantities was assessed by cluster bootstrap over runs. The number of resamples differed between analyses: two hundred for the partition of variance, whose bootstrap refits the mixed-effects model on each resample, and two thousand for each of the other three, the scale effect, the most divergent voice, and the family grouping, which resample summary statistics and are correspondingly cheaper.

The analysis pipeline was implemented with the assistance of a code-generation tool (Claude Code, Anthropic, in successive versions over the period of the work) under a cycle in which the authors specified each analysis, the tool implemented it, and the authors reviewed the implementation and its output and revised the specification where needed, occasionally editing the code directly. The analysis plan the pipeline implements was registered publicly before any of it was written, and the code is released so that its correspondence with the registered plan can be inspected. All reported figures were computed from the released artifacts and verified against them.

## Data and code availability

The analysis code is released under the Apache License 2.0, comprising the computation of the similarity matrix and of the spectral index, which depends on the reference implementation of the Vendi Score rather than reimplementing it, the per-model dissent contribution, the validity filter, the confirmatory and exploratory analyses, and the figure generator. The per-run computed indices and the model formulations from which they derive are released alongside it under a Creative Commons Attribution licence. Code and data are hosted at https://github.com/mariovegabarbas/llm-ensemble-effective-diversity and archived at Zenodo (doi: 10.5281/zenodo.21718657); the version that reproduces the figures reported here is v1.0.2. Earlier versions compute the same analyses on the same data and return the same variance components, permutation statistic and indices; they differ in expressing the fixed-effects effect sizes against the sum of squares of the analysis-of-variance table rather than against the total sum of squares of the response, which shifts those effect sizes and no other quantity. The vignette bank is held in a separate deposit, archived at Zenodo (doi: 10.5281/zenodo.21708455; Mora-Valenciano and Vega-Barbas, 2026) under CC-BY-NC 4.0, with access restricted to registered users; requests are reviewed and typically granted. The restriction is a condition set by its clinical authorship.

## Results

### The dissent regime of the panel

The panel of sixteen models produced four hundred and fifty ensemble runs over the fifteen vignettes. Of the seven thousand two hundred model-level formulations planned, forty voices were absent, leaving seven thousand one hundred and sixty formulations obtained; a further seventy-eight failed the validity filter described in the Materials and Methods and were treated as absent, leaving seven thousand and eighty-two analysed. After this filtering, three hundred and thirty-nine runs contained all sixteen valid voices, one hundred and four contained fifteen, and seven contained fourteen. Forty-seven of the obtained formulations (0.66 per cent) reached the output limit and terminated on length rather than on completion; fifteen of these survived the validity filter, 0.21 per cent of the analysed formulations.

Across the four hundred and fifty runs the spectral dissent index ranged from 0.152 to 0.233, with a mean of 0.190 and a standard deviation of 0.015. Expressed as the effective number of voices, this corresponds to a range from 1.52 to 1.91 with a mean of 1.69 (Figure 1). A nominal panel of sixteen heterogeneous models therefore behaved, on average, as the equivalent of fewer than two independent contributions per case. The condition number of the similarity matrix ranged from 284 to 711 with a mean of 454, and the largest eigenvalue accounted for a mean of 0.91 of the trace, consistent with a spectrum dominated by a single direction of consensus and a long tail of near-redundant directions. The ensemble thus operated throughout in a regime of high convergence: the models agreed with one another far more than they diverged, and the variation reported below occurs within that regime rather than across a wide spread of disagreement.

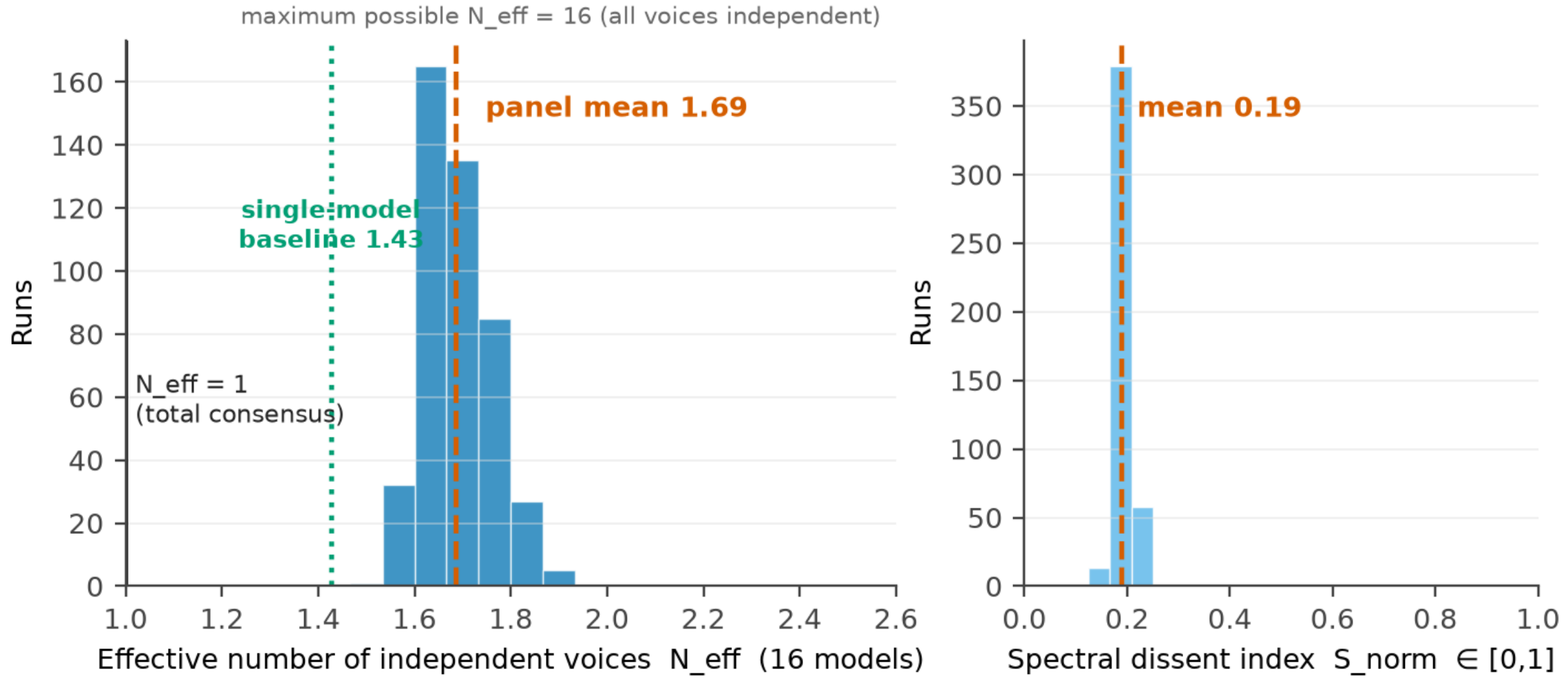

**Figure 1.** Distribution across the four hundred and fifty runs of the effective number of voices (left, panel mean 1.69) and the normalised spectral dissent index (right, mean 0.19); both concentrate near the low end of their range, with the effective number of voices far below the panel size of sixteen. The dotted line on the left marks the single-model self-consistency baseline of 1.43, the effective diversity a single model reaches on its own stochastic variation when re-run on the same case; it is a post-hoc calibration, introduced and qualified in the exploratory analyses below.

Model identifiers are reported here in their familiar form; the executed panel used version-pinned identifiers resolved against the provider catalogue on a fixed date, and the full pinned list is given in the supplementary material. The panel accordingly reflects the state of the model ecosystem at the time of execution rather than a standing set of systems.

## Confirmatory analysis

The single confirmatory hypothesis was that model identity accounts for a share of the variance in the per-model dissent contribution greater than zero. The cross-classified mixed-effects model was fitted to the seven thousand and eighty-two per-model observations with model identity and case identity as crossed random factors and no fixed-effect predictors beyond the intercept. The model converged. The estimated variance components were $1.56 \times 10^{-4}$ for the model factor, $6.58 \times 10^{-5}$ for the case factor, and $1.09 \times 10^{-4}$ for the residual.

The registered decision criterion evaluated the model-factor variance against a Monte Carlo permutation reference distribution, constructed by permuting model labels within cases over 99,999 resamples under the seed fixed in the analysis code before execution. The test statistic was the sum of squares of the model factor from a two-way analysis of variance on the case-centred response. The observed statistic was 1.0497, and none of the 99,999 permutations reached or exceeded it, placing the one-sided p-value at the floor of the procedure, $(0 + 1)/(99{,}999 + 1) = 1 \times 10^{-5}$. The preregistered sensitivity analysis, a two-way fixed-effects decomposition of the same response, gave an omega-squared of 0.464 and an eta-squared of 0.465 for the model factor, against an eta-squared of 0.194 for the case factor, agreeing with the mixed model in ordering. The confirmatory hypothesis is therefore supported: the identity of the model is a detectable structuring factor of the per-model dissent contribution, the mean distance of a model's formulation from the others in its run, read from the same similarity matrix as the spectral index but a distinct magnitude, not a decomposition of it.

The optimiser emitted a boundary warning during the fit. Inspection of the fitted model shows that both variance components have finite standard errors and that neither collapsed to zero: the model component lies approximately 2.7 standard errors from zero and the case component, which is the smaller of the two and therefore the nearer to the boundary, approximately 2.6. The warning reflects the small numerical scale of the response, whose variances fall in the order of $10^{-4}$, rather than a null component or a flat likelihood in the direction of interest. The confirmatory decision does not rest on the mixed-model fit in any case, having been made by permutation and corroborated by the fixed-effects decomposition, both of which are independent of it.

### Exploratory analyses

Six analyses are reported in this section. Four of them were specified in advance as exploratory: the partition of variance between the model and case factors, the effect of scale within families, the most divergent voice at panel scale, and the grouping of models by family. The remaining two were carried out after the fact and were not registered in any form: the single-model baseline, a calibration for the effective voice count, and the analysis of dissent by the interpretive openness of the case. Each is labelled where it appears. None of the six carries a registered directional prediction, all are reported descriptively, and none is promoted to confirmatory status. The four registered analyses are presented in the order in which they were executed, which places first the analysis most likely to complicate rather than support the confirmatory finding, followed by the two post-hoc ones. The per-model dissent contributions that these analyses decompose are shown in Figure 2.

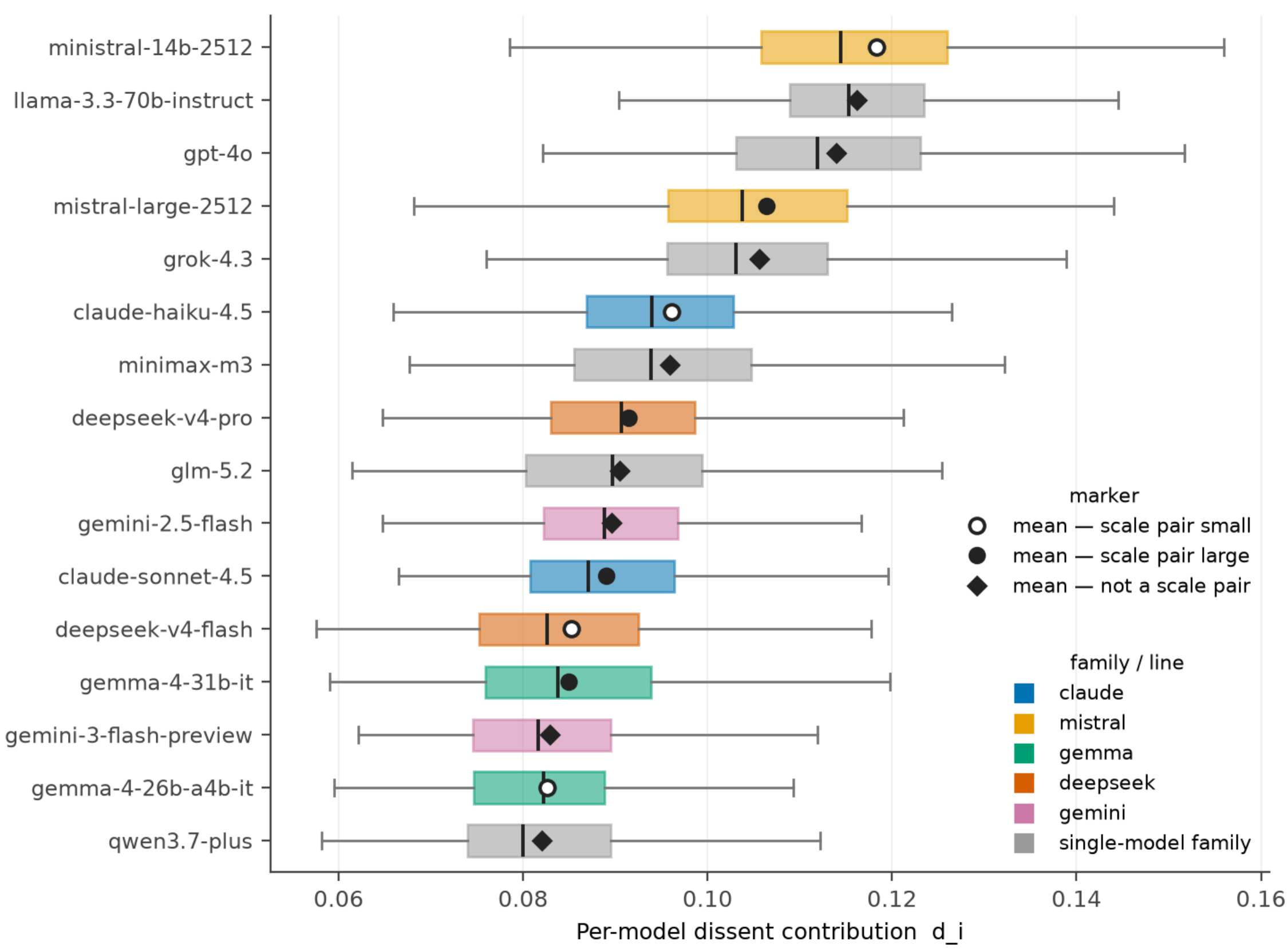

**Figure 2.** Per-model dissent contribution for the sixteen models, ordered by descending mean; each box spans the interquartile range with the median as its interior line and whiskers extending to 1.5 times the interquartile range, the overlaid marker is the mean, box colour encodes line, and marker shape encodes scale-pair role.

#### *Partition of variance between the model and case factors*

The partition of the dissent variance was characterised without any registered expectation as to which factor would account for more. The model factor accounted for 47.2 per cent of the total variance and the case factor for 19.9 per cent, with the residual absorbing the remaining 32.9 per cent. The ratio of the two structural components was 2.37. Cluster bootstrap over runs, with two hundred resamples, gave confidence intervals of [0.457, 0.489] for the model variance-partition coefficient and [0.181, 0.222] for the case coefficient, and [2.07, 2.64] for their ratio. The intervals do not overlap and the interval of the ratio does not include unity, so the model factor accounts for a larger share of the variance than the case factor, distinguishably from bootstrap noise. Excluding the one model with an appreciable shortfall of observations left the partition essentially unchanged, at 49.5 per cent and 19.7 per cent with a ratio of 2.51. The dominant feature of the partition is that the two structural factors together account for about two thirds of the variance in per-model dissent, most of it attributable to model identity, with the remaining third lying within runs.

A post-hoc check, not among the registered analyses, bears on whether this model-level structure is an artefact of the task's format. The instruction template asks each model to open its formulation by declaring a therapeutic framework from a fixed menu of ten, so the model factor can be tested against the possibility that it merely records which framework a model prefers. Refitting the two-way decomposition with the proportion of a run's other members that declared the same framework entered as a covariate leaves the model factor's eta-squared at 0.404, against 0.465 without it, so 87 per cent of it survives, while the covariate itself accounts for an eta-squared of 0.049. Separately, framework agreement is only weakly associated with lower diversity. The correlation between the modal framework's share of a run and its effective voice count is −0.37, a summary that depends on no threshold and is therefore the primary reading of the check; the gradient it summarises runs from 1.70 effective voices in the quartile of runs where framework agreement is lowest to 1.64 in the quartile where it is highest. Agreement on framework contributes to the measured similarity but does not account for the model-level structure, which is largely independent of framework choice. The detail is given in S9 of the supplementary material.

### *Scale within families*

Four pairs of models differing in scale within the same line were declared in advance: Claude Haiku 4.5 against Claude Sonnet 4.5, Ministral 14B against Mistral Large, Gemma 4 26B against Gemma 4 31B, and DeepSeek V4 Flash against DeepSeek V4 Pro. Within each pair the paired difference in dissent contribution was distinguishable from bootstrap noise over two thousand resamples, but the direction of the difference was not shared across pairs (Figure 3). In the Claude and Mistral pairs the smaller model separated further from the consensus than the larger, by 0.0070 with an interval of [−0.0082, −0.0058] and by 0.0120 with an interval of [−0.0133, −0.0107] respectively. In the Gemma and DeepSeek pairs the larger model separated further, by 0.0024 with an interval of [0.0016, 0.0033] and by 0.0061 with an interval of [0.0048, 0.0073] respectively. The Claude and Mistral comparisons rest on all four hundred and fifty runs; the validity filter removed some voices from the Gemma and DeepSeek pairs, leaving four hundred and forty-seven and four hundred and forty-two paired runs respectively. With four pairs and no shared direction, this is reported as an observation about these four pairs rather than as evidence bearing on a general effect of scale, and no aggregate statistic was computed across pairs.

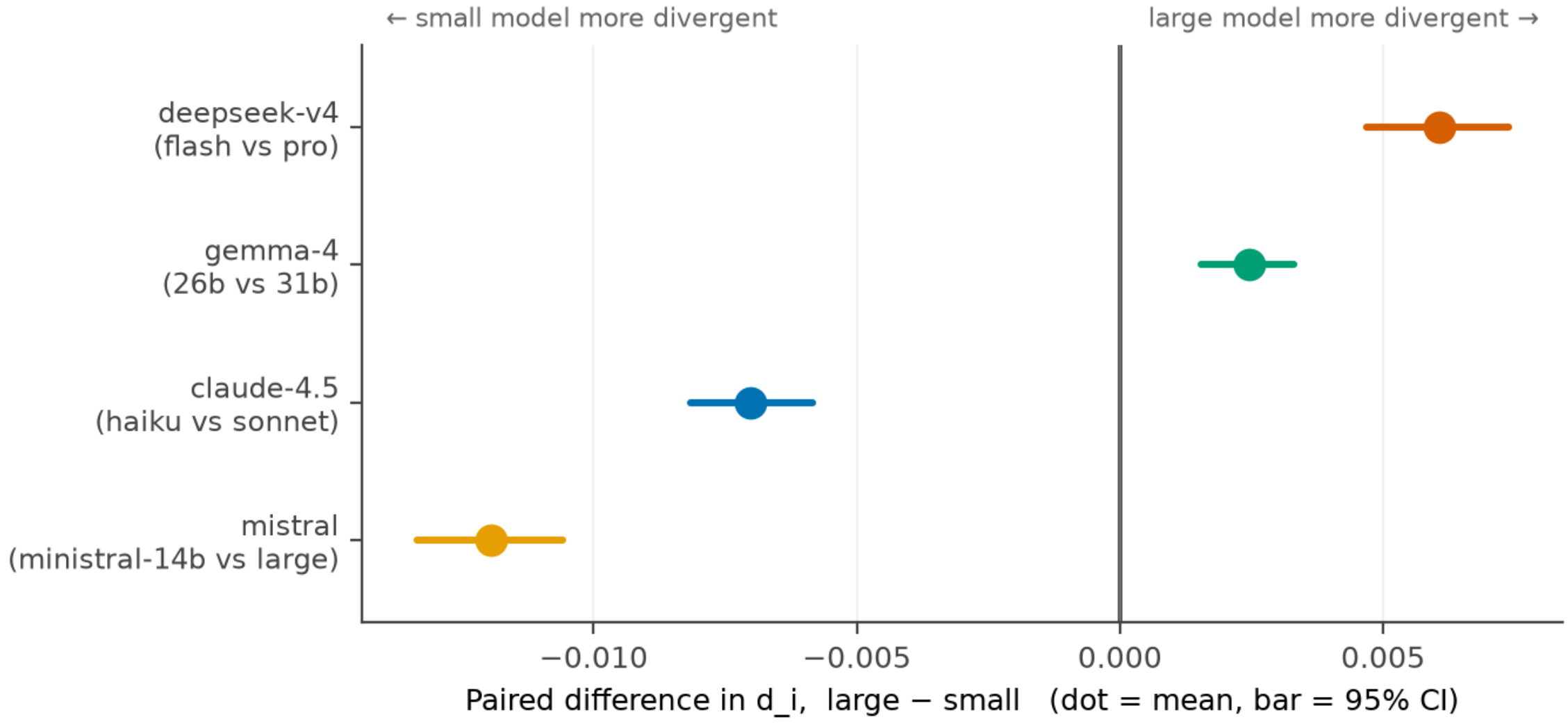

**Figure 3.** Paired within-family difference in dissent contribution (large minus small) for the four scale pairs, with 95 per cent confidence intervals; each difference is distinguishable from zero but its sign is not shared across pairs, the smaller model being more divergent in the Claude and Mistral pairs and the larger in the Gemma and DeepSeek pairs.

### *The most divergent voice at panel scale*

The voice separating most from the consensus was identified in each run among its valid voices. Under interchangeability each model would occupy that position in about 6.25 per cent of runs, an expectation of between twenty-five and twenty-nine runs depending on how many runs the model contributed. Three models stood appreciably above that expectation, with bootstrap intervals excluding it: Ministral 14B in 35.1 per cent of runs, Llama 3.3 70B in 28.4 per cent, and GPT-4o in 20.0 per cent. Grok 4.3 sat at the boundary, in 6.2 per cent of runs with an interval spanning the chance expectation. The remaining models fell below it, and four, Claude Sonnet 4.5, Gemini 3 Flash, Gemma 4 26B and Gemma 4 31B, were never the most divergent voice in any run. Gemini 2.5 Flash, which held the position in a tenth of runs before the validity filter, does so in a single run once its degenerate outputs are excluded. Normalising the counts by the number of valid runs each model contributed does not alter this picture. The complete distribution across the sixteen models is given in the supplementary material.

The identity of the most divergent voice was not stable across cases. Ministral 14B was modal in seven vignettes and Llama 3.3 70B in six, with GPT-4o modal in the remaining two. This last figure is worth setting against an earlier three-model exercise on the same vignette bank, in which GPT-4o was the most divergent voice in sixty-seven per cent of runs and modal in thirteen of the fifteen vignettes. In the sixteen-model panel it falls to third place, holding the position in 20.0 per cent of runs and modal in two vignettes, behind two models that were absent from the smaller ensemble. The position of most divergent voice therefore appears to depend on the composition of the ensemble rather than to be an intrinsic attribute of a model, since the same model changes rank as the surrounding panel changes.

### *Grouping by model family*

The sixteen models belong to ten families, six of them represented by a single model and none by more than four; the five lines with two members are the only ones that permit any within-line comparison. Mean dissent contribution ranged from 0.082 to 0.118 across models. The full per-model contributions, with their dispersion and the number of runs contributing to each, are given in the supplementary material. Dispersion within lines, at 0.0069, was smaller than dispersion between lines, at 0.0146, giving a ratio of 0.469 with a bootstrap interval of [0.435, 0.502]. The ratio falls below unity with an interval excluding it, consistent with a grouping effect by which two models of the same line resemble one another more than two models of different lines; the interval is narrow because the bootstrap resamples runs, but the comparison itself rests on only five two-member lines.

The grouping is more homogeneous than the pre-filter data suggested. All five two-member lines have members that lie close together, the within-line difference ranging from 0.0024 for Gemma to 0.0120 for Mistral; the two Gemini generations, which before the validity filter appeared as far apart as distinct families, differ by only 0.0067 once their degenerate outputs are removed. No generational step within a line matches the typical difference between lines. Any reading of a family signature rests here on two models per line at most, and on a single model for the majority of families, so the observed grouping is reported with that limitation attached, and the conclusion is confined to the five two-member lines represented. Compared with published profiles of behavioural metacognition across model families (Abtahi et al., 2026), the convergence is structural and weak: both that work and the present measurements find that family membership carries information, but the constructs differ, and no correspondence can be drawn between a metacognitive profile and a value of the dissent index.

### *A single-model baseline*

This analysis was not preregistered and is reported as a calibration for the effective voice count. It asks what the index returns when the panel is replaced by a single model, so that all variation is that one model's stochastic variation. For each model and each vignette, sixteen of the model's own valid formulations of that vignette were drawn at random, their Gram matrix was formed, and the effective number of voices was computed; the draw was repeated thirty times per pair under a fixed seed, over the two hundred and thirty-nine of two hundred and forty model-vignette pairs with at least sixteen valid formulations. A single model re-run in this way produced, on average, the equivalent of 1.43 distinct formulations, against the 1.69 of the sixteen-model panel.

The baseline overstates intra-model diversity, and does so in a known direction. The sixteen formulations drawn for it span different presentation orders of the framework list, whereas the sixteen voices of a real run share one order; the baseline therefore carries order-induced variation that the real measurement does not, and is an upper bound on the diversity a single model would show at fixed order. The gap between the panel and the baseline, 0.26 effective voices, is correspondingly a lower bound on what the heterogeneity of the panel contributes over a single model's own variability. A fixed-order baseline, which would remove the confound, is not available: each order was run only three times, too few to form a sixteen-voice matrix. The full distribution is given in the supplementary material.

### *Dissent and the interpretive openness of the case*

The vignette bank was stratified along a second dimension besides clinical picture: the interpretive openness of the case, in three levels, six vignettes for which a shared framework reads the case ineffectively, two for which a shared framework reads it effectively, and seven that genuinely admit distinct framings. That stratification bears directly on what the index registers, since a measure of dissent among readings would be expected to return more dissent on the cases built to admit several. The check was not among the registered analyses and was carried out after the fact.

It does not come out. The seven vignettes designed to admit distinct framings returned a mean of 1.675 effective voices, against 1.695 for the six on which a shared framework reads the case ineffectively and 1.705 for the two on which it reads the case effectively. The difference between the first two strata is 0.020 effective voices in the direction opposite to the one the construct predicts, with a bootstrap interval over runs of [0.006, 0.033]. That interval treats the thirty runs of a vignette as independent; resampling vignettes instead, which is the unit the stratification applies to, gives intervals that include zero for every contrast. The stratum of two vignettes is reported for completeness and supports no comparison.

The marginal difference is a composition effect. Dissent varies far more with clinical picture than with openness: the four pictures span 0.136 effective voices, from 1.632 for trauma to 1.768 for depression, against 0.030 across the openness strata. Depression, the picture with the most dissent, is over-represented among the ineffective-framework vignettes, and computing the same contrast within each picture reverses its sign in three of the four, averaging to −0.0002 at equal weight. A second and independent check agrees: the share of a run's members declaring the modal framework does not distinguish the stratum designed to admit distinct framings from the one designed not to, at 0.691 against 0.697 with an interval of [−0.047, 0.037].

Two readings are compatible with this. The index may not track interpretive openness, or the openness labels may not identify it. The labels were a design premise of the bank and were never validated independently: the four therapists who reviewed the bank rated plausibility and sufficiency, not openness. The design does not separate the two, and on this material the dissent the ensemble produces is organised by the clinical content of the case rather than by the range of readings the case was built to admit. What the analysis establishes positively is therefore narrower than what it was designed to test: the quantity varies systematically on this material, but with the clinical content of the case rather than with the property the bank was stratified on.

## Discussion

### Structure without explanatory categories

The confirmatory test establishes that model identity structures how far a formulation falls from the ensemble consensus. It does not say which property of a model accounts for the position it occupies, and the exploratory analyses were designed to ask that.

Neither of the two categories ordinarily used to group language models supports, on this design, a claim that it predicts a model's position. Scale within a family points in different directions across pairs, and family membership shows a grouping effect that rests on too few lines to establish more.

Scale does not. Within each of the four scale pairs, the paired difference in dissent contribution was distinguishable from bootstrap noise, so scale is not inert; but the direction of that difference was not shared across the pairs. In two of them the smaller model separated further from the consensus, in the other two the larger did. The differences are individually clear and collectively incoherent. This is not the pattern of an effect too small to detect; it is the pattern of an effect whose sign depends on the family. The nearest published evidence, MEDLEY-BENCH, a benchmark of behavioural metacognition over a much larger panel (Abtahi et al., 2026), reports scale effects that are similarly dissociated and non-monotonic, with smaller models frequently matching or exceeding larger ones. Two instruments measuring different constructs converge on the conclusion that scale is not a summary variable for model behaviour. Evidence from the mental-health domain itself points the same way: within a single model family, a small fine-tuned model matched much larger ones in the higher-data categories of an emotional-safety classification task (Pinzuti et al., 2025).

Family membership groups more consistently than the pre-filter data suggested. All five two-member lines have members that sit close together, and the one apparent exception, the two Gemini generations, turns out to have been an artefact of degenerate outputs concentrated in one of them. Within-line dispersion is about half of between-line dispersion. The evidential basis, however, is five paired comparisons: the bootstrap interval is narrow because it resamples runs rather than lines, and the reading would have been different had a single line behaved otherwise. Family membership is therefore consistent with a grouping effect on this panel, but the design does not support a claim that it predicts a model's position.

The picture that emerges is of structure at the level of the individual model. Each model occupies a characteristic position that its size does not predict and that its family recovers only on the thin evidence of five paired lines. Model-level signatures of this kind have also been reported in a clinical-safety evaluation across multiple language models (Santos et al., 2026). For a field accustomed to reasoning about models in terms of family and scale, that recommends measuring a model's behaviour rather than inferring it from its label.

### The most divergent voice belongs to the ensemble

The voice that separates most from the consensus is not a property of a model but of the company it keeps. Under interchangeability each of the sixteen would occupy that position in about 6 per cent of runs. Three stood appreciably above that expectation, Ministral 14B in 35.1 per cent of runs, Llama 3.3 70B in 28.4 per cent and GPT-4o in 20.0 per cent, and four never occupied it in any of the four hundred and fifty runs. The position is therefore far from uniformly distributed, and to that extent it does carry information about models.

What it does not carry is information that survives a change of panel. In an earlier three-model exercise on the same material, GPT-4o was modal in thirteen of the fifteen vignettes. In the sixteen-model panel it falls to third, modal in two, behind two models that were absent from the smaller ensemble. Nothing changed in GPT-4o between the two exercises; what changed was the company. The identity of the most divergent voice is a relational quantity, defined against the particular set of members present, and the same applies to its complement: a model that never diverges most in this panel may well do so in another.

This is one of the study's principal results, not a corollary of the others, and it has a direct consequence for any system that surfaces a dissenting voice to a human reader. What such a system displays is determined by the composition of the panel, not by a stable attribute of the model displayed. Changing one member can change which voice is presented as the outlier without anything having changed in the model that was presented before. An interface built on that notion should treat the dissenting voice as a statement about the current ensemble, should make the composition of that ensemble visible to whoever is reading it, and should not report the outlier as a property the model carries from one deployment to another.

### What fewer than two effective voices means

Sixteen models, deliberately drawn from ten families and two scales, produced on average the semantic diversity of fewer than two independent contributions. Heterogeneity of provenance did not translate into heterogeneity of output. This extends to a heterogeneous multi-model panel in an interpretive clinical domain a convergence that has been reported for creative and open-ended tasks, where models have been found to be homogeneous in their outputs across providers (Wenger and Kenett, 2025), where the collapse has been traced to the composition of post-training data (Karouzos et al., 2026), and where added agents have been found to yield diminishing marginal diversity in multi-agent arrangements (Chen et al., 2026). The present measurements place that phenomenon in a domain selected for the genuine plurality of defensible readings it admits, which is the setting in which an ensemble would be expected to diverge if it were going to.

The immediate design implication is that adding models to an ensemble does not add voices in proportion. If the purpose of an ensemble is to widen the space of readings offered to a human decision-maker, the number of models is a poor proxy for how far that space has been widened, and the effective voice count is the quantity that should be reported. An ensemble of five well-separated models may offer more than an ensemble of twenty that converge, and only a measure invariant to panel size can compare the two, which is why the intensive formulation matters for this purpose. A single-model baseline sharpens the point. Re-running one model on the same case already yields the equivalent of about 1.43 distinct formulations, so the sixteen-model panel's 1.69 stands only about a quarter of a voice above one model's own stochastic variation. Because that baseline mixes presentation orders a real run does not, it overstates the single-model diversity, so the quarter-voice gap is a lower bound on what the panel's heterogeneity contributes over a single model: at least that much, and plausibly more.

Why the convergence is so pronounced cannot be fully settled here, but the three candidate explanations are no longer on an equal footing. One is a property of the task: the models are asked to select a framework from a closed menu and to write from it, so that those choosing the same framework begin from the same conceptual apparatus and produce texts that resemble one another for reasons partly terminological. A post-hoc analysis, reported in S9 of the supplementary material and not preregistered, places a bound on this explanation: agreement on framework is associated with only about six hundredths of an effective voice, so the closed menu does not account for the convergence in any essential way. The other two explanations remain open and cannot be separated with this design. The convergence may be a property of the models, trained on overlapping corpora with overlapping objectives and perhaps converged on a shared way of writing about clinical material, or a property of the material, if the vignettes admit fewer defensible readings than their construction intended. The openness check reported below bears on this second explanation without resolving it: the material does not behave as its stratification anticipated, but whether that is because the vignettes admit fewer readings than intended or because the index does not register the difference is precisely what the design cannot separate. Distinguishing the two remaining explanations requires varying the material against a fixed panel, which is the complement of the design used here.

### What the dissent tracks

This study measures dispersion on a task for which no correctness criterion exists, so what the quantity registers there cannot be established by validating it against error, and has to be characterised against the properties of the material instead. The case bank was stratified along a dimension besides clinical picture, the interpretive openness of the case, and that stratification permits a check on what the index registers: a measure of dissent among readings should return more dissent on the cases built to admit several readings. It does not. The seven vignettes designed to admit distinct framings returned slightly fewer effective voices than the rest, a difference that is small and in the direction opposite to the one the construct predicts. Computing the same contrast within each clinical picture reverses its sign in three of the four and averages to −0.0002 effective voices at equal weight, which is what identifies the marginal difference as a composition effect. A second and independent instrument agrees: the share of a run's members converging on one declared therapeutic framework does not distinguish the stratum designed to admit distinct framings from the stratum designed not to.

What the dissent does track, on this material, is clinical content. The four clinical pictures span 0.136 effective voices, from trauma at the low end to depression at the high, against 0.030 across the openness strata. Two readings of that result remain open and this design does not separate them: the index may not register interpretive openness, or the openness labels may not identify it. The labels were a premise of the bank's construction and were never validated independently, so the check is a joint one, on the measure and on the labels at the same time, and its negative outcome does not apportion between them. It is tempting to read the ordering by clinical picture as evidence that dissent tracks whether a presentation has a canonical way in, trauma affording one and depression less so. That is a conjecture these data suggest and do not establish; testing it requires material built to vary that property against a fixed panel.

### Where the measure sits

The programme within which this study sits treats confidence as infrastructure, in the sense of the distinction between confidence, the functional relation a person maintains with a system they rely on without choosing, and trust, the deliberate acceptance of risk in an agent (Luhmann, 1988). Instruments in that programme measure the confidence a human experiences toward an opaque system. The quantity characterised here is on the other side of that relation: not what an operator feels, but how far the members of a system fall from one another. That is the sense in which the two belong to one programme, and it is the origin of this study rather than an application it delivers. Whether measured dispersion could calibrate perceived confidence is a further question, and nothing reported here licenses an affirmative answer. The measure does not register the property of the case for which the material was stratified, and no study with human supervisors in the loop has been run. The regulatory requirement of meaningful human oversight of high-risk systems is real (European Parliament and Council of the European Union, 2024), and an auditable account of when a system does not speak with one voice would bear on it; but the step from a measure of dispersion to an instrument that tells a supervisor when to intervene is not taken here, and this study does not support taking it.

### Limitations

The material is a single bank of fifteen vignettes in a single clinical domain, and its validation is weak in psychometric terms: agreement among the four raters was poor on both dimensions, so the bank is established as plausible and representative characterisation material rather than as a psychometrically validated instrument. Results obtained on it should not be assumed to transfer to other domains or to other case material without replication. A further limitation concerns the openness stratification itself. The three levels of interpretive openness were assigned by the clinician who wrote the vignettes, as a premise of the bank's construction, and were never validated independently: the four therapists who reviewed the bank rated the clinical plausibility and the clinical sufficiency of each vignette and the representativeness of the bank as a whole, and were not asked whether a vignette admits one defensible reading or several. The check reported in the Results is therefore a joint one, bearing on the index and on the labels at the same time, and its negative outcome does not apportion between them: an index that fails to register interpretive openness and a set of labels that fails to identify it produce the same result here. Separating the two requires case material whose interpretive openness has been established by raters independent of its author, which this bank does not provide. The bank and the formulations are in Spanish, a language that occupies a minor fraction of the corpora on which these models are trained, so both the magnitude of the dissent and its structure could differ in other languages; the study does not test this.

The task itself constrains what the index can measure. Models are asked to select a therapeutic framework from a closed list of ten and to write from it, so the menu bounds the space of formulations they can produce, and the index quantifies dissent under structured rather than free formulation. A post-hoc analysis, reported in S9 of the supplementary material and not among the registered analyses, bears on two distinct questions that the closed menu raises. The first is whether the model-level structure is a record of framework preference rather than of anything else: conditioning the model factor on the share of a run that declared the same framework leaves 87 per cent of its eta-squared intact. The covariate is in fact an attribute of the case and not of the model, since model identity accounts for 5.1 per cent of its variance and case identity for 46.6 per cent, and it overlaps accordingly with the case factor, which falls further than the model factor does, from 0.194 to 0.159. The second question is whether the convergence itself is an artefact of the format, and the relevant quantity is a different one: the effective number of voices is only about six hundredths lower in the most framework-homogeneous runs than in the most heterogeneous, and even in the quartile where framework agreement is highest it stands at 1.64 rather than falling towards one. The convergence is therefore not attributable in any essential way to the format of the task, even though the menu still bounds the formulation space and free formulation might return different values. The framework declared in each formulation was recorded. The instruction template, inherited unchanged from the earlier exercise so that generation conditions would be comparable, informs the model that its reading will be compared with others; making the ensemble setting explicit is a defensible design choice, but its effect on how models position themselves was not examined.

The panel is sixteen models observed at one point in time, with identifiers pinned to dated versions. Most families are represented by a single model and none by more than two in the lines that permit comparison, so family-level readings rest on very few cases. One model contributed 8.4 per cent fewer runs than the rest owing to a transient provider failure, though the reported partitions are unchanged by its exclusion. One model was removed from the registered panel before execution because its reasoning could not be disabled, and the output limit was raised above the registered value because the registered value truncated two models; both departures are documented and were adopted before any data were generated.

A further limitation is of a different kind: the index measures dispersion in semantic space and says nothing about quality. A model that separates from the consensus may be contributing a defensible minority reading or may be producing a worse formulation, and nothing in these measurements distinguishes the two. The same holds for the ensemble as a whole: a high effective voice count indicates that the models disagree, not that the disagreement is informative. A post-hoc analysis, not among the registered analyses, bears on one particular way in which a formulation can separate without differing in content. Normalising the embeddings removes magnitude but not the direction correlated with extension, so departing from the typical length of a run contributes to the dissent contribution irrespective of what is said. The association is with the absolute deviation and not with signed length: writing at length does not move a voice from the consensus, departing from the run in either direction does, and the three models that most often occupy the position of most divergent voice sit on opposite sides of the length distribution. Extension is not, however, an external variable that could simply be controlled for. Model identity accounts for 78.3 per cent of the variance in absolute length deviation, so conditioning on it is partly conditioning on the factor being measured, and the resulting estimate depends on the order in which the terms enter, ranging from 46 to 100 per cent of the model factor surviving; the difficulty of separating a diversity measure from the capability of the models it is computed over has been made in a different setting (Kim, 2026). What the analysis does rule out is length as the explanation of the finding on clinical picture: mean length differs across the four pictures by 4.4 per cent of the shortest, against a within-picture dispersion 7.4 times that spread, and the ordering by length does not follow the ordering by dissent.

About a third of the variance in per-model dissent lies within runs and is left unexplained: the study establishes that model identity structures the dissent, not that it accounts for most of it. That residual is not decomposed here, so what varies from one run of a model to another on the same case, whether decoding stochasticity, the presentation order or something the design does not isolate, remains open.

## Conclusions

The identity of the model is a detectable structuring factor of the per-model dissent contribution, of how far a model's formulation falls from the ensemble consensus. That was the single hypothesis registered before any data were generated, and it is supported: the variance component attributable to model identity is greater than zero under the registered criterion, with no permutation of the ninety-nine thousand nine hundred and ninety-nine drawn reaching the observed statistic, and the registered sensitivity analysis agrees in ordering. Which model produces a formulation matters for that distance, and it matters beyond what the case alone accounts for.

That structure is fine structure. A panel of sixteen models drawn from ten families, presented with cases stratified by clinical picture and by interpretive openness, behaved on average as the equivalent of 1.69 distinct formulations, and the largest eigenvalue of the similarity matrix accounted for about nine tenths of its trace. What the confirmatory test detects is organisation within a narrow band of disagreement, not the organisation of a wide field of it. Both halves of that statement matter: model identity carries information, and it carries it within a narrow band. For anyone assembling an ensemble, this has a practical consequence: the effective diversity a panel yields is not predictable from its size, from the properties of its members or from their provenance, so it is a quantity to be measured on the task at hand rather than assumed from the composition of the panel.

What the quantity registers, on this material, is the clinical content of the case rather than the interpretive openness the bank was stratified on, and two explanations of that remain open which this design does not separate. Distinguishing them requires case material whose interpretive openness has been established independently of its author, varied against a fixed panel: the complement of the design used here, and one the present study makes both necessary and tractable, the contribution of the task format having already been bounded and found small.

## Ethics Statement

The study involved no human participants as research subjects and no patient data. The clinical vignettes are synthetic, written for this purpose, and describe no real person. Four practising therapists were invited to review the vignette bank and contributed their assessments through an anonymous survey; participation was voluntary and no personal or identifying data were collected. Under these conditions the review did not constitute research on human subjects and was not submitted for ethics committee approval.

## Author Contributions

MV-B: Conceptualization, Methodology, Software, Formal Analysis, Investigation, Data curation, Visualization, Writing – original draft, Writing – review and editing, Project administration, Funding acquisition, Supervision. LM-V: Conceptualization, Resources, Investigation, Validation, Writing – review and editing. IP: Conceptualization, Methodology, Supervision, Writing – review and editing. FA: Conceptualization, Methodology, Validation, Supervision, Writing – review and editing. FS: Conceptualization, Supervision, Writing – review and editing.

## Funding

This work was supported by the internal research project SIPPIA (Sistema de Interacción Segura, Privada y Personalizada para Aplicaciones Telemáticas basado en Tecnología Ubicua e Inteligencia Artificial Generativa), Universidad Politécnica de Madrid, reference OTT-UPM RP2409230067.

## Conflict of Interest

The authors declare that the research was conducted in the absence of any commercial or financial relationships that could be construed as a potential conflict of interest.

## Generative AI Statement

The authors used AI-based tools in the preparation of this manuscript. Large language models (Claude, Anthropic, in successive versions over the period of the work) were used to support and enhance the writing of the manuscript, from results supplied and verified by the authors, Scite was used to assist with the literature search and citation analysis, and Grammarly was used for English proofreading. The authors independently verified all cited sources. The use of a code-generation tool in implementing the analysis pipeline is described in the Materials and Methods. These tools did not contribute to the conceptual content, the study design, the choice of measure, or the scientific conclusions, all of which are the authors' own. The sixteen language models evaluated in this study are its object of investigation and not tools used in its preparation. The authors take full responsibility for the content of the manuscript.

## Supplementary Material

This supplementary material collects the tables to which the main text refers and the additional detail required to reproduce the analyses. All values are computed from the released artifacts and reported at the precision used in the analysis.

### S1. Panel composition

This table gives the version-pinned identifier of each of the sixteen models, its family and its line, its membership in a within-family scale pair and role within that pair, and, of its four hundred and fifty runs, the number in which it contributed a valid voice, the number excluded as degenerate by the validity filter, and the number absent. It is the complete pinned list to which the Materials and Methods (panel of models) and the Results (grouping by model family) refer with "the full pinned list is given in the supplementary material".

| Pinned identifier | Family | Line | Scale pair | Role | Valid | Degen. | Absent |
|---|---|---|---|---|---|---|---|
| anthropic/claude-4.5-haiku-20251001 | anthropic | claude | claude-4.5 | small | 450 | 0 | 0 |
| anthropic/claude-4.5-sonnet-20250929 | anthropic | claude | claude-4.5 | large | 450 | 0 | 0 |
| deepseek/deepseek-v4-flash-20260423 | deepseek | deepseek | deepseek-v4 | small | 442 | 8 | 0 |
| deepseek/deepseek-v4-pro-20260423 | deepseek | deepseek | deepseek-v4 | large | 450 | 0 | 0 |
| google/gemini-2.5-flash | google | gemini | | | 401 | 49 | 0 |
| google/gemini-3-flash-preview-20251217 | google | gemini | | | 450 | 0 | 0 |
| google/gemma-4-26b-a4b-it-20260403 | google | gemma | gemma-4 | small | 450 | 0 | 0 |
| google/gemma-4-31b-it-20260402 | google | gemma | gemma-4 | large | 447 | 3 | 0 |
| meta-llama/llama-3.3-70b-instruct | meta | meta | | | 440 | 10 | 0 |
| minimax/minimax-m3-20260531 | minimax | minimax | | | 405 | 7 | 38 |
| mistralai/ministral-14b-2512 | mistral | mistral | mistral | small | 450 | 0 | 0 |
| mistralai/mistral-large-2512 | mistral | mistral | mistral | large | 450 | 0 | 0 |
| openai/gpt-4o | openai | openai | | | 450 | 0 | 0 |
| qwen/qwen3.7-plus-20260602 | qwen | qwen | | | 450 | 0 | 0 |
| x-ai/grok-4.3-20260430 | xai | xai | | | 449 | 1 | 0 |
| z-ai/glm-5.2-20260616 | zai | zai | | | 448 | 0 | 2 |

The family column follows the provider taxonomy, under which the four Google models form a single family; the line column follows the finer taxonomy, under which the two Gemini generations and the two Gemma models form distinct lines. The two Gemini generations share a line but are not a declared scale pair. The identifiers were resolved against the provider catalogue at the time of the run of 22 July 2026 and were frozen before any data were generated; each refers to a specific dated snapshot rather than to a floating alias. Two identifiers (openai/gpt-4o and google/gemini-2.5-flash) carry no date suffix because the provider exposes them as already-stable snapshots. Excluded voices are of two kinds: outputs that failed the validity filter, concentrated in Gemini 2.5 Flash, and voices absent because the provider returned nothing usable, concentrated in MiniMax M3; both are detailed in S4 and S7.

### S2. Per-model dissent contribution

This table summarises the per-model dissent contribution across the runs each model contributed, ordered by descending mean to match the order of Figure 2. It supports the per-model dissent reported in the Results (grouping by model family) and the family analysis of the Discussion.

| **Model** | **Mean d_i** | **SD** | **IQR (Q1–Q3)** | **n** |
|---|---|---|---|---|
| Ministral 14B | 0.1183 | 0.0184 | 0.1058–0.1260 | 450 |
| Llama 3.3 70B | 0.1162 | 0.0105 | 0.1089–0.1234 | 440 |
| GPT-4o | 0.1139 | 0.0143 | 0.1031–0.1231 | 450 |
| Mistral Large | 0.1064 | 0.0151 | 0.0957–0.1151 | 450 |
| Grok 4.3 | 0.1056 | 0.0138 | 0.0957–0.1130 | 449 |
| Claude Haiku 4.5 | 0.0961 | 0.0127 | 0.0869–0.1028 | 450 |
| MiniMax M3 | 0.0959 | 0.0153 | 0.0855–0.1048 | 405 |
| DeepSeek V4 Pro | 0.0914 | 0.0126 | 0.0830–0.0986 | 450 |
| GLM 5.2 | 0.0904 | 0.0126 | 0.0803–0.0994 | 448 |
| Gemini 2.5 Flash | 0.0896 | 0.0103 | 0.0822–0.0968 | 401 |
| Claude Sonnet 4.5 | 0.0890 | 0.0117 | 0.0808–0.0964 | 450 |
| DeepSeek V4 Flash | 0.0853 | 0.0144 | 0.0753–0.0925 | 442 |
| Gemma 4 31B | 0.0850 | 0.0114 | 0.0760–0.0938 | 447 |
| Gemini 3 Flash | 0.0829 | 0.0111 | 0.0745–0.0895 | 450 |
| Gemma 4 26B | 0.0826 | 0.0104 | 0.0747–0.0888 | 450 |
| Qwen 3.7 Plus | 0.0820 | 0.0114 | 0.0740–0.0895 | 450 |

The standard deviations are uniform across models, between 0.010 and 0.018, and the mean and interquartile range agree closely for every model. The right skew that inflated two models' standard deviations before the validity filter was contributed entirely by the degenerate outputs, now excluded.

### S3. Most divergent voice, full distribution

This table gives, for each of the sixteen models, the number of runs in which it was the most divergent voice, that count as a percentage of the four hundred and fifty runs, its ninety-five per cent cluster-bootstrap interval, and the rate normalised by the number of runs the model contributed. It is the complete distribution to which the Results (the most divergent voice at panel scale) refer with "the complete distribution across the sixteen models is given in the supplementary material". Under interchangeability each model would occupy the position in about 6.25 per cent of runs, an expectation of between twenty-five and twenty-nine of the four hundred and fifty, depending on the number of runs to which the model contributed a valid voice. The most divergent voice is the model with the largest per-model dissent in a run; the quantity is continuous and no ties arise, and a first-occurrence rule would break one if it did.

| Model | Runs as most divergent | % of 450 | 95% bootstrap CI | Rate by valid runs |
|---|---|---|---|---|
| Ministral 14B | 158 | 35.11 | [138, 178] | 0.3511 |
| Llama 3.3 70B | 128 | 28.44 | [110, 147] | 0.2909 |
| GPT-4o | 90 | 20.00 | [74, 107] | 0.2000 |
| Grok 4.3 | 28 | 6.22 | [18, 38] | 0.0624 |
| Mistral Large | 23 | 5.11 | [15, 33] | 0.0511 |
| MiniMax M3 | 7 | 1.56 | [3, 13] | 0.0173 |
| DeepSeek V4 Pro | 6 | 1.33 | [2, 11] | 0.0133 |
| DeepSeek V4 Flash | 3 | 0.67 | | 0.0068 |
| Claude Haiku 4.5 | 2 | 0.44 | | 0.0044 |
| Qwen 3.7 Plus | 2 | 0.44 | | 0.0044 |
| GLM 5.2 | 2 | 0.44 | | 0.0045 |
| Gemini 2.5 Flash | 1 | 0.22 | | 0.0025 |
| Claude Sonnet 4.5 | 0 | 0.00 | ≤ 3 | 0.0000 |
| Gemini 3 Flash | 0 | 0.00 | ≤ 3 | 0.0000 |
| Gemma 4 26B | 0 | 0.00 | ≤ 3 | 0.0000 |
| Gemma 4 31B | 0 | 0.00 | ≤ 3 | 0.0000 |

Cluster-bootstrap intervals are reported only for the seven models observed in the position at least five times. Below that count the percentile interval is uninformative, with a lower bound of zero even where events occurred, so the count is reported without an interval; for the four models never observed, a rule-of-three calculation bounds the rate above by three runs in four hundred and fifty, about 0.67 per cent. The normalised rate divides by the number of valid runs each model contributed, which is below four hundred and fifty for the seven models with excluded voices (S1); the reduced denominators do not alter the ordering. Four models, Claude Sonnet 4.5, Gemini 3 Flash, Gemma 4 26B and Gemma 4 31B, were never the most divergent voice.

### S4. Missing data by model and vignette

This section expands the summary given in the Materials and Methods (handling of incomplete runs) and the Discussion (limitations). Forty of the four hundred and fifty runs were computed from fifteen voices rather than sixteen. Only two models account for the absences: MiniMax M3 in thirty-eight runs and GLM 5.2 in two. The absences fall in a contiguous window of the run sequence and are consistent with a transient provider instability rather than with any property of the cases.

| Vignette | Absent voice and count |
|---|---|
| Vignette 3 | MiniMax M3 ×6 |
| Vignette 5 | MiniMax M3 ×9 |
| Vignette 6 | MiniMax M3 ×14 |
| Vignette 7 | MiniMax M3 ×7 |
| Vignette 8 | GLM 5.2 ×1 |

| Vignette | Absent voice and count |
|---|---|
| Vignette 10 | GLM 5.2 ×1, MiniMax M3 ×1 |
| Vignette 15 | MiniMax M3 ×1 |

The remaining eight vignettes had no absence in any run: Vignettes 1, 2, 4, 9, 11, 12, 13 and 14.

One further matter of data handling is recorded here. Forty-seven of the seven thousand one hundred and sixty obtained formulations (0.66 per cent) reached the 4096-token output limit; fifteen of these survived the validity filter, 0.21 per cent of the analysed formulations. The seventy-eight voices excluded by that filter are broken down by criterion and by model in S7.

### S5. Raw (unnormalised) entropy

This section reports the descriptive statistics of the unnormalised von Neumann entropy S, in nats, across the four hundred and fifty runs. The main text (Materials and Methods, the dissent index) reports the entropy normalised by log n and its exponential, the effective number of voices; the raw entropy is retained here. It relates to the reported quantities by S equal to the natural logarithm of the effective number of voices, equivalently S_norm times log n, where n is the number of voices in the run.

- **Minimum:** 0.4202
- **Maximum:** 0.6465
- **Mean:** 0.5223
- **Standard deviation:** 0.0411
- **Median:** 0.5186

### S6. Reproducibility details

This section collects the seeds, input hashes, resample counts and library versions needed to reproduce the analyses.

- **Order-plan seed:** 42
- **Prompt template SHA-256:** 596518f4018ea981669f756b9002245f709898adab0ea49c37afc30a1270b396
- **Vignette bank SHA-256:** ffd5d46b1c0f3a7efc105ab58ce22cc0c2c38c20d5eae03206270b2e0b52c1c2
- **Confirmatory analysis seed:** 20260722
- **Confirmatory permutation resamples:** 99,999 (Monte Carlo permutation of model labels within cases; p-value floor $1\times10^{-5}$)
- **Cluster-bootstrap resamples, variance partition (RQ5):** 200 (each resample refits the cross-classified mixed-effects model)
- **Cluster-bootstrap resamples, scale effect (RQ3), most divergent voice (RQ4) and family grouping (RQ2):** 2,000 each
- **Mixed-effects fit:** statsmodels 0.14.6 (MixedLM), on numpy 2.4.1 and scipy 1.17.1
- **Embeddings:** OpenAI text-embedding-3-large, 3072 dimensions, via the openai client 2.16.0

The four library versions listed above are those recorded at the time of the analysis. The environment was not captured in full when the study was run, so the versions of the remaining dependencies, among them pandas and the reference implementation of the Vendi Score, are not on record and cannot be reconstructed after the fact. The released code declares minimum versions in `requirements.txt`; that file is a dependency declaration and not a lock file, and it is not offered as a record of the environment that produced the figures reported here.

Reproduction has two layers, and they differ in what they require. The analyses reported in this paper run from the published per-run indices and per-model dissent contributions, and reproduce the reported figures from them. The computation of the index itself begins one step earlier, from the embeddings of the formulations; those embeddings are not published, and regenerating them requires re-embedding the released texts, as the repository documents. That step was verified separately: recomputing the effective number of voices from the original embeddings, under the frozen definition of the index, reproduces the published value in all four hundred and fifty runs to within $5 \times 10^{-11}$.

**S7. Degenerate formulations**

This section expands the summary given in the Materials and Methods (validity of formulations) and is the breakdown to which S4 refers with "the seventy-eight voices excluded by that filter are broken down by criterion and by model in S7". The validity filter removed seventy-eight of the seven thousand one hundred and sixty formulations obtained, each then treated as an absent voice under the same missing-voice rule as the provider absences. Classified by the first of the seven criteria to fire, in the order in which they are numbered in the validity criteria document, sixty were provider errors, flagged by a finish reason of error and, in forty-nine cases, by zero output tokens as well; fourteen were shorter than the 150-word threshold; and four were in a language other than Spanish, two of them the MiniMax M3 outputs whose character set changed part way through generation. The character-repetition, n-gram-loop and non-alphabetic criteria fired only as redundant confirmations on responses already caught by these three, on three, ten and three responses respectively, and excluded nothing the first three criteria did not; the withdrawn framework-declaration criterion is discussed in the validity criteria document.

The failures were concentrated in a few models, and their nature differed by model. Gemini 2.5 Flash, DeepSeek V4 Flash and Gemma 4 31B contributed only provider errors; Llama 3.3 70B, MiniMax M3 and Grok 4.3 contributed short or wrong-language outputs. The remaining ten models produced no degenerate formulation.

| **Model** | **Provider error** | **Too short** | **Wrong language** | **Total** |
|---|---|---|---|---|
| Gemini 2.5 Flash | 49 | 0 | 0 | 49 |
| Llama 3.3 70B | 0 | 8 | 2 | 10 |
| DeepSeek V4 Flash | 8 | 0 | 0 | 8 |
| MiniMax M3 | 0 | 5 | 2 | 7 |
| Gemma 4 31B | 3 | 0 | 0 | 3 |
| Grok 4.3 | 0 | 1 | 0 | 1 |
| Total | 60 | 14 | 4 | 78 |

At the level of the run, seventy-five of the four hundred and fifty runs contained at least one degenerate voice, seventy-two of them one and three of them two; four of these runs also carried a provider-absent voice. Treating the degenerate voices as absent, no run fell below fourteen valid voices of sixteen: sixty-eight of the seventy-five contaminated runs retained fifteen valid voices and seven retained fourteen, so every run stayed above the floor at which the index remains interpretable.

### S8. Single-model baseline distribution

This section reports the distribution of the single-model baseline described in the Results (a single-model baseline). For each model and each vignette, sixteen of that model's valid formulations of the vignette were drawn at random, their Gram matrix was formed, and the effective number of voices was computed; the draw was repeated thirty times per pair under seed 20260728. Of the two hundred and forty model-vignette pairs, two hundred and thirty-nine had at least sixteen valid formulations and entered the baseline; one, MiniMax M3 on Vignette 6, had only thirteen valid formulations and was excluded. Across the pairs the effective number of voices had a mean of 1.427, a standard deviation of 0.090, a minimum of 1.194 and a maximum of 1.718. The panel of sixteen models averaged 1.687 effective voices across the four hundred and fifty runs, so the gap between the panel and the single-model baseline is 0.260 effective voices. Because the sixteen formulations drawn for each baseline pair span different presentation orders of the framework list, whereas the sixteen voices of a real run share one order, the baseline carries order-induced variation the real measurement does not; it is therefore an upper bound on the diversity a single model would show at fixed order, and the gap is a lower bound on what the heterogeneity of the panel contributes, as set out in the Results.

### S9. Framework choice: conditioned partition and homogeneity (post-hoc)

The instruction template requires each formulation to open by declaring a therapeutic framework from a fixed menu of ten. The declared framework was recovered for 99.75 per cent of the seven thousand and eighty-two analysed formulations by an extended parser whose patterns were defined from the template and the observed forms of declaration, without reference to the dissent contribution or the identity of the model; the parser, its extraction rates and the descriptive distribution of the declaration are reported in S10. The two analyses below are post-hoc and were not preregistered.

**Conditioned partition.** For each observation a framework-agreement covariate was formed as the proportion of the other valid members of the run that declared the same framework; it has a mean of 0.58 and a correlation of −0.37 with the per-model dissent contribution. Entering this covariate in the two-way decomposition of the per-model dissent leaves the model factor's eta-squared at 0.404, against 0.465 without it, so 87 per cent of the model factor survives; the covariate accounts for an eta-squared of 0.049 and the case factor falls from 0.194 to 0.159. The cross-classified mixed model agrees: the model variance component falls from $1.56 \times 10^{-4}$ to $1.39 \times 10^{-4}$, an eleven per cent reduction, with a covariate coefficient of −0.017 (standard error 0.0005). Because the choice of framework is largely case-driven, the covariate overlaps mainly with the case factor rather than with the model factor.

**Homogeneity partition.** Each run was given a modal-framework share, the fraction of its valid members declaring the most common framework, with a median of 0.75. A run was classified as framework-homogeneous, a threshold fixed before the effective voice count was tabulated by subset, if this share was at least one half. The three hundred and seventy homogeneous runs averaged 1.684 effective voices and the eighty heterogeneous runs 1.701, a difference of 0.017. Across the four quartiles of the modal share the mean effective voice count was 1.702, 1.735, 1.675 and 1.638, giving a difference of 0.064 between the extreme quartiles, and the correlation between the modal share and the effective voice count was

−0.37. These three readings of the same gradient are reported together because they differ, and the manuscript quotes the largest of them: the contrast between extreme quartiles is the steepest slope the gradient attains, not the most favourable reading of it, while the split at the pre-declared threshold gives 0.017 and the correlation, which depends on no threshold at all, is the primary summary. The threshold of one half was fixed before the effective voice count was tabulated by subset, and its result is reported here for that reason, even though the analysis as a whole is post-hoc. Framework agreement is thus associated with lower effective diversity, but the association is small: even where nearly all models declare the same framework, the effective voice count stays near 1.64, close to the panel mean of 1.69 and far below the panel size.

**S10. Declared framework: descriptive distribution and parser extension (post-hoc)**

This section records the descriptive distribution of the declared framework across the panel and the extension of the parser that recovers it. Like S9, it is post-hoc and was not preregistered, and it is reported descriptively only: nothing is claimed here about bias, about how these frequencies relate to the prevalence of each framework in the clinical literature, or about why some frameworks are declared more often than others. The choice of framework is in large part determined by the case, the median modal-framework share of a run being 0.75 (S9), so the distribution below reflects the vignette bank as much as it reflects the models.

**Parser extension.** The original parser recovers the declaration in the form the template prescribes. Two variant forms and a set of orthographic wrappings escaped it, and its patterns were extended to cover them. All patterns were defined from the instruction template and from the observed surface forms of declaration, without reference to the per-model dissent contribution or to the identity of the model that produced the formulation, and the extension only recovers formulations the original parser left unresolved: it reclassifies none that the original already mapped. Two patterns were added at the structural level: a generalisation of the template lead-in, searched over the first six hundred characters, which admits the declaration when it appears as a Markdown heading or without the opening clause of the template; and a labelled form in which the framework follows a bold or heading label. Three were added at the level of the match between the captured text and the menu: the boundary after a canonical label was relaxed to admit a comma, period, semicolon or colon and not only a space, while continuing to reject a letter continuation, so that it cannot map a longer word onto a shorter label; the generic descriptive lead-ins "basado en" and "centrado en" are stripped and the remainder retried, with the direct match attempted first; and the already-sanctioned alias between the adjectival and nominal forms of the narrative label is applied to the head of a compound declaration.

The extension raises the proportion of the seven thousand and eighty-two valid formulations in which a declaration is found structurally from 99.52 to 99.89 per cent, and the proportion mapped to one of the ten menu frameworks from 98.76 to 99.75 per cent. Structural failures fall from thirty-four to eight and unmapped captures from eighty-eight to eighteen. The lowest per-model mapped rate after the extension is 97.53 per cent, for MiniMax M3, followed by 99.33 per cent for Ministral 14B; every other model is at or above 99.56 per cent. Of the eighteen residual formulations, eight are MiniMax M3 responses whose stored content begins part way through the declaration, a data-quality artefact of the truncated responses rather than a third form of declaration; these were deliberately not recovered by hand, since doing so would be a model-specific adjustment of the kind the blindness condition excludes. The remaining ten carry an abbreviated or compound declaration whose primary label falls outside the menu and would require a semantic judgement to assign.

**Declared framework by model.** The table gives the declared framework of each of the seven thousand and eighty-two valid formulations, by model. The frameworks are given in the order of the menu and abbreviated: CB cognitive-behavioural, HE humanistic-existential, PD psychodynamic-psychoanalytic, INT integrative, MF mindfulness-based or third-wave, ATT attachment, TRA somatic/sensorimotor trauma, SST structural-strategic systemic, TGS transgenerational systemic (Bowen), NAR narrative therapy. The column n.e. counts the formulations in which no declaration could be mapped, n is the number of valid formulations of that model, and H is the Shannon entropy in bits of its distribution over the mapped frameworks, whose maximum for a menu of ten is $\log_2 10 = 3.322$.

| Model | CB | HE | PD | INT | MF | ATT | TRA | SST | TGS | NAR | n.e. | n | H |
|---|---|---|---|---|---|---|---|---|---|---|---|---|---|
| Claude Haiku 4.5 | 0 | 0 | 80 | 12 | 1 | 77 | 128 | 62 | 81 | 7 | 2 | 450 | 2.491 |
| Claude Sonnet 4.5 | 11 | 0 | 106 | 74 | 0 | 6 | 120 | 80 | 48 | 5 | 0 | 450 | 2.502 |
| DeepSeek V4 Flash | 0 | 0 | 98 | 62 | 0 | 12 | 144 | 78 | 47 | 0 | 1 | 442 | 2.335 |
| DeepSeek V4 Pro | 0 | 0 | 102 | 58 | 0 | 27 | 110 | 76 | 77 | 0 | 0 | 450 | 2.476 |
| Gemini 2.5 Flash | 0 | 1 | 143 | 34 | 0 | 17 | 109 | 71 | 26 | 0 | 0 | 401 | 2.256 |
| Gemini 3 Flash | 0 | 0 | 86 | 11 | 0 | 42 | 134 | 96 | 63 | 18 | 0 | 450 | 2.485 |
| Gemma 4 26B | 3 | 0 | 100 | 4 | 0 | 69 | 120 | 88 | 43 | 22 | 1 | 450 | 2.514 |
| Gemma 4 31B | 0 | 2 | 104 | 2 | 0 | 101 | 120 | 65 | 45 | 8 | 0 | 447 | 2.395 |
| GLM 5.2 | 9 | 0 | 93 | 42 | 0 | 41 | 120 | 86 | 57 | 0 | 0 | 448 | 2.565 |
| GPT-4o | 0 | 4 | 106 | 0 | 2 | 71 | 121 | 81 | 65 | 0 | 0 | 450 | 2.365 |
| Grok 4.3 | 3 | 1 | 58 | 59 | 7 | 84 | 123 | 68 | 46 | 0 | 0 | 449 | 2.641 |
| Llama 3.3 70B | 2 | 0 | 92 | 176 | 0 | 3 | 91 | 48 | 27 | 0 | 1 | 440 | 2.153 |
| MiniMax M3 | 6 | 1 | 52 | 26 | 2 | 127 | 54 | 37 | 89 | 1 | 10 | 405 | 2.541 |
| Ministral 14B | 0 | 0 | 9 | 281 | 0 | 2 | 54 | 80 | 21 | 0 | 3 | 450 | 1.589 |
| Mistral Large | 0 | 0 | 28 | 187 | 0 | 52 | 82 | 66 | 35 | 0 | 0 | 450 | 2.276 |
| Qwen 3.7 Plus | 2 | 0 | 64 | 57 | 1 | 93 | 127 | 30 | 68 | 8 | 0 | 450 | 2.593 |

The per-model entropy has a mean of 2.386 bits, a minimum of 1.589 for Ministral 14B, which declares the integrative framework in two hundred and eighty-one of its four hundred and fifty formulations, and a maximum of 2.641 for Grok 4.3.

**Panel aggregate.** Over the seven thousand and sixty-four mapped formulations, the declared framework is distributed across the ten menu options as follows: somatic/sensorimotor trauma 1,757 (24.9 per cent), psychodynamic-psychoanalytic 1,321 (18.7 per cent), structural-strategic systemic 1,112 (15.7 per cent), integrative 1,085 (15.4 per cent), transgenerational systemic 838 (11.9 per cent), attachment 824 (11.7 per cent), narrative therapy 69 (1.0 per cent), cognitive-behavioural 36 (0.5 per cent), mindfulness-based or third-wave 13 (0.2 per cent), and humanistic-existential 9 (0.1 per cent). The entropy of the aggregate distribution is 2.646 bits of the maximum 3.322, or 0.797 when normalised by that maximum.

**Near-empty frameworks.** No menu option is entirely absent at panel level, but four are near-empty, and their scarcity is shared across the panel rather than confined to a few models. Humanistic-existential is declared in 0.1 per cent of the mapped formulations and is never declared by eleven of the sixteen models; mindfulness-based or third-wave in 0.2 per cent, never declared by eleven; cognitive-behavioural in 0.5 per cent, never declared by nine; and narrative therapy in 1.0 per cent, never declared by nine. Of the remaining six options, five are declared at least once by every model and the integrative framework by fifteen of the sixteen.

**S11. Continuity between the determinant-based and the spectral formulation (registered check)**

The registration specifies, as an internal coherence check of the index and not as a hypothesis test, that the three reference cases of the earlier three-model pilot be recomputed under the spectral formulation and shown to preserve the ordering produced by the earlier determinant-based one. The three cases were recomputed from their stored similarity matrices with the frozen definition used throughout this study.

| Reference case | Determinant-based index | Determinant | S_norm | N_eff |
|---|---|---|---|---|
| REF-001 | 0.3151 | 0.031273 | 0.2821 | 1.363 |
| REF-003 | 0.2977 | 0.026379 | 0.2605 | 1.331 |
| REF-002 | 0.2829 | 0.022648 | 0.2463 | 1.311 |

The determinant-based index orders the three cases REF-001, REF-003, REF-002 from most to least dissent, and the spectral index orders them identically. The two formulations therefore operationalise the same construct on this material. The check involves three cases and three voices and is reported as a property of the index computed within this project, not as evidence about models.